\documentclass[times,preprint,10pt]{elsarticle}
\usepackage{amsmath,amssymb,booktabs,array,multirow,graphicx,xspace,microtype,float}
\usepackage{setspace,algpseudocode,pdflscape,placeins}
\floatstyle{ruled}
\newfloat{algorithm}{tbp}{loa}
\floatname{algorithm}{Algorithm}
\makeatletter
\let\floatc@ruled\@makecaption
\makeatother
\usepackage[hidelinks]{hyperref}
\biboptions{numbers,sort&compress}
\newcommand{\method}{PCLM\xspace}

\newcommand{\R}{\mathbb{R}}
\newcommand{\norm}[1]{\operatorname{norm}\!\left(#1\right)}
\newcommand{\thd}[1]{\textbf{#1}}

\newcommand{\tablesize}{\singlespacing\normalsize\setlength{\tabcolsep}{4pt}\renewcommand{\arraystretch}{1.12}}

\makeatletter
\let\ps@pprintTitle\ps@plain
\makeatother
\hypersetup{pdftitle={PCLM: Small-target localization with frozen CLIP via prototype contrast and local magnification},pdfauthor={Zhipeng Ye, Feng Jiang, Qiufeng Wang, Hao Li},pdfsubject={Support-conditioned small-target localization},pdfcreator={LaTeX with Elsevier elsarticle}}
\begin{document}
\begin{frontmatter}
\title{PCLM: Small-target localization with frozen CLIP via prototype contrast and local magnification}
\author[1]{Zhipeng Ye}\ead{zhipengye@nustti.edu.cn}
\author[1]{Feng Jiang\corref{cor1}}\ead{jf@nustti.edu.cn}
\author[2]{Qiufeng Wang}\ead{qiufeng.wang@xjtlu.edu.cn}
\author[3]{Hao Li}\ead{lihao@arizona.edu}
\address[1]{Taizhou Institute of Science and Technology, Nanjing University of Science and Technology, Taizhou 225300, Jiangsu, China}
\address[2]{Department of Intelligence Science, Xi'an Jiaotong-Liverpool University, Suzhou 215123, Jiangsu, China}
\address[3]{School of Computer Science and Technology, University of Arizona, Tucson 85705, AZ, USA}
\cortext[cor1]{Corresponding author.}
\begin{abstract}
Small targets occupy few patches in a vision-language encoder, so spatial features often mix object appearance with surrounding content. We propose Prototype Contrast and Local Magnification (PCLM), a support-conditioned localization method that uses a frozen CLIP encoder. Five masked support images per class define foreground and background prototypes through equally weighted regional features. Their difference provides a shared scoring direction for query patches, explicitly comparing target evidence with the demonstrated background. Nine overlapping query windows are enlarged and encoded independently to sample small targets more densely. Reprojection and coverage averaging combine their scores into a continuous localization map. The class direction occupies 2 KiB regardless of support count and transfers unchanged across datasets with mapped categories. On 5,047 small-target queries from VOC, COCO, ADE20K and Oxford-IIIT Pets, PCLM achieves higher mean pixel AP than every evaluated text-conditioned localization baseline on each dataset under our evaluation protocol. Gains over the strongest scene-dataset baselines range from 5.63 to 13.23 percentage points. At comparable measured latency, local magnification improves scene small-target AP by 4.51--5.68 points over whole-canvas enlargement. Factorial experiments show that prototype contrast increases the benefit of local observation, including under matched image-coordinate filtering. Support-budget experiments show that additional examples refine category estimation without increasing representation size or query-time scoring cost.
\end{abstract}
\begin{keyword}
\interlinepenalty=10000
Small-target localization \sep Prototype contrast \sep Local magnification \sep Frozen~CLIP \sep Visual supports
\end{keyword}
\end{frontmatter}
\thispagestyle{plain}
\section{Introduction}
Small objects are difficult to localize when a vision transformer observes an entire scene at a fixed input size. A target may occupy only a few patches, and each patch can mix object appearance with surrounding content. Localization therefore requires both enough spatial samples of the object and a score that distinguishes it from its surroundings. Contrastive Language--Image Pre-training (CLIP)~\cite{radford2021clip} provides a useful starting point: its pretrained features encode visual categories, while its patch tokens retain the spatial information needed to locate them.

\begin{figure}[!tb]
\centering\includegraphics[width=\linewidth]{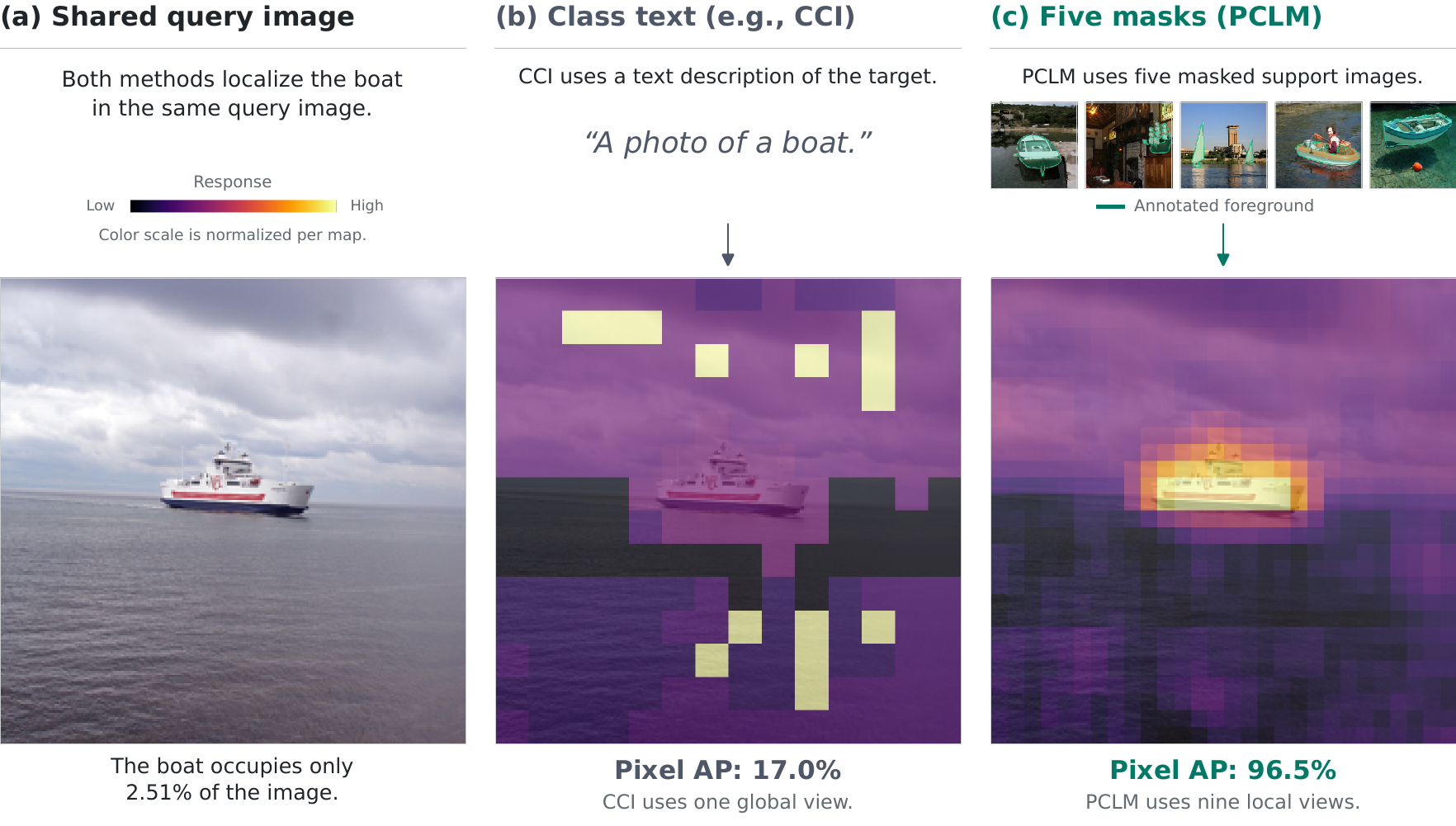}
\caption{A small-boat example from VOC. CCI~\cite{agarwal2026cci} uses class text and one global view (17.0\% pixel AP); PCLM uses five masked supports from bank 11 and nine local views (96.5\% AP). The target occupies 2.51\% of the canvas. AP uses original scores over known foreground and background pixels; each heatmap is scaled separately for display.}
\label{fig:intro_comparison}
\end{figure}

Existing CLIP localization methods commonly specify a target with class text and extract a spatial response from image--text matching. Visual examples offer another way to specify a category. A foreground mask identifies which parts of each example belong to the target and which belong to its background. Figure~\ref{fig:intro_comparison} illustrates this difference on a small boat: CCI~\cite{agarwal2026cci} uses the sentence ``A photo of a boat.'', while PCLM uses five masked support images. In this example, CCI highlights several surrounding regions, while PCLM concentrates its response on the boat.

We study category-conditioned localization with a frozen CLIP encoder. Given a few masked support images and a query containing the category, the method produces a continuous heatmap of the target region. All instances of the category contribute to that region. In our evaluation, a small target occupies less than 5\% of the query canvas after these instances are combined.

We propose \method, which combines \emph{prototype contrast} and \emph{local magnification}. Five masked supports define a foreground prototype and a background prototype. Each support contributes one normalized feature vector to each prototype, so images with larger masks do not dominate the category representation. A query patch is scored by its similarity to the foreground prototype minus its similarity to the background prototype. This difference measures how strongly the patch resembles the demonstrated target relative to the demonstrated background.

Local magnification provides more patch samples of the target. PCLM extracts nine overlapping windows from the query canvas, enlarges each window and encodes it independently. Every view uses the same foreground--background direction. The score maps are smoothed, mapped back to their window locations and averaged wherever windows overlap. The design keeps category construction separate from query observation: more supports refine the prototypes, while more views increase spatial sampling and encoding cost. A class direction occupies 2 KiB regardless of the number of supports.

We evaluate this combination on VOC, COCO, ADE20K and Oxford-IIIT Pets, reusing VOC-derived prototypes for the mapped categories in every dataset. The experiments examine three aspects of the design: how prototype contrast changes the benefit of local views; how local re-encoding compares with whole-image enlargement under matched filtering or comparable latency; and how accuracy changes with support count, view count and encoder choice. Across 5,047 small-target queries, the combined method improves localization on all four datasets.

Our contributions are:
\begin{enumerate}
\item \textbf{A shared foreground--background score for local views.} PCLM combines equally weighted support-region prototypes with independently enlarged query windows. Their shared contrast direction and coverage averaging produce a dense localization map using frozen CLIP features.
\item \textbf{Evidence that contrast and local observation work together.} Factorial experiments show that contrast increases local-observation gains on the three scene datasets. Resolution comparisons at comparable measured latency and experiments with the same image-coordinate filter establish the benefit of local re-encoding.
\item \textbf{A reusable category representation with separate support and view budgets.} VOC-derived prototypes transfer unchanged across mapped categories in the four datasets. Support-count, view-count, backbone and mask-boundary studies characterize accuracy and stability while the per-class storage remains fixed as supports increase.
\item \textbf{Strong small-target localization results.} With five masked supports per class, PCLM achieves higher mean small-target AP than every evaluated text-conditioned baseline on all four datasets. Gains over the strongest baseline on each scene dataset range from 5.63 to 13.23 percentage points.
\end{enumerate}

\section{Related work}
\subsection{Visual localization and spatial explanation}
CLIP learns image--text alignment through global contrastive supervision~\cite{radford2021clip}, while its transformer patch states retain spatial information~\cite{dosovitskiy2021vit}. More generally, attention rollout and attention flow aggregate token interactions across transformer layers to estimate input relevance~\cite{abnar2020attention}. Chefer et al. propagate relevance through attention layers and skip connections~\cite{chefer2021transformer}. These approaches trace the information used by a transformer prediction.

Text-conditioned localization methods use spatial features to explain image--text matching. Grad-CAM weights spatial activations with gradients of a selected output~\cite{selvaraju2017gradcam}, and CLIPCAM applies this principle to text-guided object and action localization~\cite{hsia2022clipcam}. Grad-ECLIP combines channel sensitivity with spatial weights on token features~\cite{zhao2026gradeclip}, whereas LeGrad aggregates positive attention gradients across transformer layers~\cite{bousselham2025legrad}. gScoreCAM combines gradient-based channel selection with model responses to masked images~\cite{chen2022gscorecam}. These methods use the sensitivity of an image--text score to identify relevant locations.

CCI measures how image--text similarity changes when groups of patch features are masked in attention, producing a heatmap from normalized similarity drops~\cite{agarwal2026cci}. CLIP Surgery changes attention and removes redundant feature components to obtain more localized image--text similarities~\cite{li2025clipsurgery}. Both derive spatial evidence from a text-specified category. \method specifies the category with masked visual examples and scores each query patch by its relative similarity to foreground and background prototypes. This gives one fixed category direction that can be applied to independently encoded local views. Table~\ref{tab:positioning} compares the target information and spatial readouts of the evaluated localization methods.

Related work on weakly supervised object localization studies how classifiers reveal object extent. Weakly supervised foreground learning learns foreground masks from pseudo boxes to guide localization and detection~\cite{zhang2023wsfl}. SeCM combines semantic matching between original and locally shuffled images with CLIP-guided cross-attention for transformer-based localization~\cite{cao2025secm}. Lee et al. improve localization at inference by binarizing penultimate features and their corresponding classifier weights~\cite{lee2024inference}. \method uses supplied support masks to summarize foreground and background regions, then applies their contrast to independently encoded local views with a frozen encoder.

% Portrait table: column widths sum to \linewidth after inter-column spacing.
\begin{table}[tbp]
\centering
\caption{Target information and spatial readouts in the localization comparison. Parentheses give publication venues and years. Scores describe the evaluated outputs.}
\label{tab:positioning}
\begingroup
\renewcommand{\baselinestretch}{1}\normalsize
\renewcommand{\arraystretch}{1}
\setlength{\tabcolsep}{4.5pt}
% Each body cell has the same height, with vertically centered content.
\newcommand{\positioncell}[1]{\parbox[c][48pt][c]{\linewidth}{\raggedright #1\par}}
\begin{tabular}{@{}>{\raggedright\arraybackslash}p{\dimexpr 0.35\linewidth-2.10\tabcolsep\relax}>{\raggedright\arraybackslash}p{\dimexpr 0.13\linewidth-0.78\tabcolsep\relax}>{\raggedright\arraybackslash}p{\dimexpr 0.25\linewidth-1.5\tabcolsep\relax}>{\raggedright\arraybackslash}p{\dimexpr 0.27\linewidth-1.62\tabcolsep\relax}@{}}
\toprule
\textbf{Method (venue/year)} & \textbf{Target} & \textbf{Spatial computation} & \textbf{Heatmap score}\\
\midrule
\positioncell{CCI (CVPR'26)~\cite{agarwal2026cci}} & \positioncell{Class text} & \positioncell{Final-block cluster masking} & \positioncell{Normalized cosine-similarity drop}\\
\positioncell{Grad-ECLIP (TPAMI'26)~\cite{zhao2026gradeclip}} & \positioncell{Class text} & \positioncell{Channel and spatial weighting} & \positioncell{Matching relevance}\\
\positioncell{CLIP Surgery (PR'25)~\cite{li2025clipsurgery}} & \positioncell{Class text} & \positioncell{Attention and feature surgery} & \positioncell{Dense image--text similarity}\\
\positioncell{Grad-CAM (IJCV'20)~\cite{selvaraju2017gradcam}} & \positioncell{Class text} & \positioncell{Activation--gradient aggregation} & \positioncell{Positive class activation}\\
\positioncell{gScoreCAM (ACCV'22)~\cite{chen2022gscorecam}} & \positioncell{Class text} & \positioncell{Gradient-selected mask scoring} & \positioncell{Score-weighted activation}\\
\positioncell{LeGrad (ICCV'25)~\cite{bousselham2025legrad}} & \positioncell{Class text} & \positioncell{Positive attention gradients} & \positioncell{Feature-formation sensitivity}\\
\midrule
\positioncell{\method (this work)} & \positioncell{Five masks} & \positioncell{Shared prototype contrast; local magnification} & \positioncell{Foreground--background cosine margin}\\
\bottomrule
\end{tabular}
\endgroup
\end{table}

\subsection{Visual supports and prototype contrast}
Few-shot classifiers differ in how they compare queries with labeled supports. Matching Networks predict labels through attention over support examples~\cite{vinyals2016matching}. Prototypical Networks average support embeddings into class centers and classify queries by their distances to those centers~\cite{snell2017prototypical}, while Relation Networks learn a nonlinear comparison between support and query embeddings~\cite{sung2018relation}.

With frozen CLIP, Tip-Adapter retrieves cached support image features and labels to supplement text-based classification scores without training~\cite{zhang2022tip}. Descriptions provide another source of category information: Menon and Vondrick use language-model-generated visual descriptors for image classification~\cite{menon2023description}. IDEA pairs individual support images with descriptions, combining image-based and description-based similarities with CLIP zero-shot scores~\cite{ye2026idea}. These approaches produce image-level class scores from example retrieval or textual descriptions.

Learned adaptation changes the prompts or feature readout. CoOp optimizes text context vectors~\cite{zhou2022coop}; CoCoOp generates image-conditioned text prompts~\cite{zhou2022cocoop}; and MaPLe learns coupled prompts across the visual and language branches~\cite{khattak2023maple}. CLIP-Adapter learns a bottleneck adapter and blends its output with the original features~\cite{gao2024clipadapter}. These methods fit additional parameters for image classification.

\method uses support masks to form two region-level centers for each category. Each image contributes a normalized foreground direction and a normalized background direction, with equal weight across supports. Their difference directly scores spatial query features, without fitting prompts or a comparison module. The same direction is reused across query images and local views, and its storage remains fixed as more supports are added. This design turns the demonstrated foreground and background into a shared spatial scoring rule.

\subsection{Local observations and spatial sampling}
Multi-scale recognition can combine features inside a network or change the images presented to it. Feature Pyramid Networks combine high-level and high-resolution features through top-down and lateral connections for object detection~\cite{lin2017fpn}. FeatUp learns feature upsampling through multi-view consistency, restoring spatial detail in pretrained representations~\cite{fu2024featup}. Scaling on Scales ($S^2$) runs a frozen encoder on sub-images from several image scales, pools their feature maps to a common spatial size and concatenates the maps across scales~\cite{shi2025scales}. \method obtains denser observations by independently re-encoding enlarged windows. It applies the same foreground--background direction within each view and combines spatial scores after reprojection. Its final representation is a class-specific localization map.

Slicing Aided Hyper Inference (SAHI) enlarges overlapping image slices, runs an object detector on each slice and merges the resulting boxes with non-maximum suppression~\cite{akyon2022sahi}. \method uses local enlargement to allocate more patch samples to a small target. Its nine windows are taken from the same query canvas and encoded independently, so each window changes both the sampling density and the surrounding context. A shared prototype direction scores all views, and the reprojected scores are averaged over the windows covering each location. This fusion combines continuous localization evidence while accounting for unequal spatial coverage.
\FloatBarrier

% Method overview and notation.
\section{Prototype contrast and local magnification}
\label{sec:method}
\subsection{Task and overview}
A query consists of an RGB image $I_q$ and a category $c$ known to be present. A support set $\mathcal S_c=\{(I_i,M_i)\}_{i=1}^{K}$ demonstrates the category, with $K=5$ in the main configuration. Each support mask labels category foreground as 1, known background as 0 and ignored pixels as 255. The output is a real-valued heatmap $H_c\in\R^{224\times224}$ that ranks query locations by their correspondence to the demonstrated category. All instances of the requested category belong to the same foreground map.

\begin{figure}[t]
\centering\includegraphics[width=\linewidth]{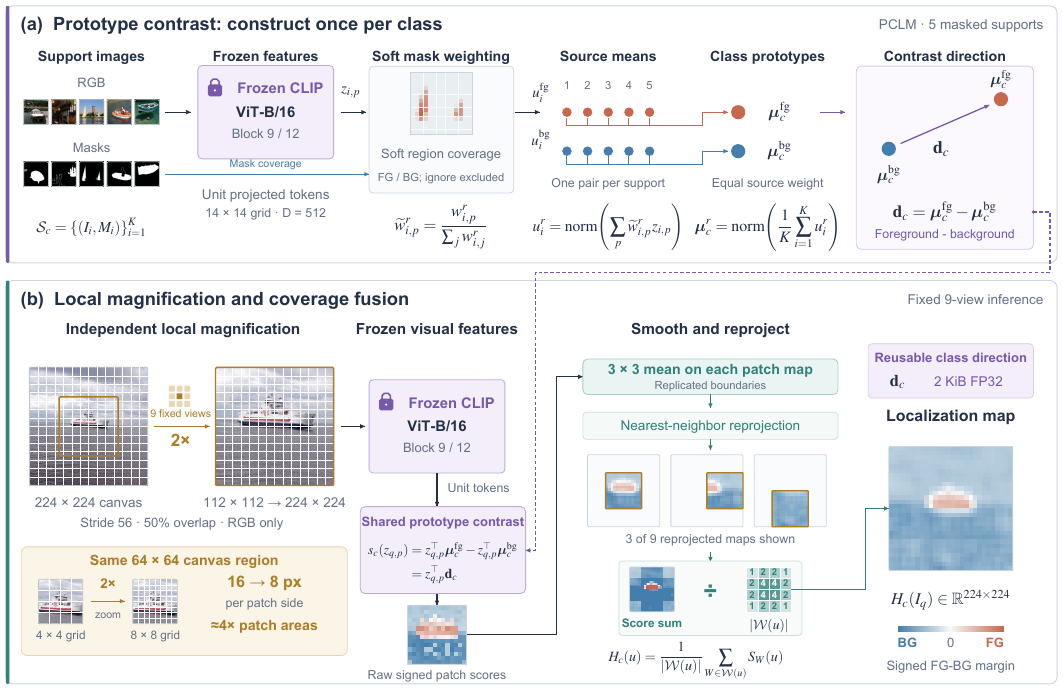}
\caption{PCLM. (a) Five support images and their masks define foreground and background prototypes, with equal weight for each image. Their difference gives the reusable class direction $\mathbf d_c$. (b) Nine overlapping query windows are enlarged and independently encoded. The shared direction produces signed patch scores, followed by mean filtering, nearest-neighbor reprojection and coverage averaging. Local magnification halves the nominal patch footprint on the query canvas and allocates approximately four times as many patch areas to a visible target. The boat example illustrates the observation and fusion stages.}
\label{fig:method}
\end{figure}

PCLM has two stages (Fig.~\ref{fig:method}). First, it pools foreground and background features from each support image, combines them into two class prototypes and stores their difference. Second, it independently encodes nine enlarged query windows, scores every patch with that shared difference vector and averages the overlapping predictions. The class representation is built once and reused for subsequent queries.

\subsection{Frozen spatial features}
We use OpenAI CLIP ViT-B/16 in evaluation mode with all weights frozen. Each complete support RGB image or query window is resized to $224\times224$ by bicubic interpolation, scaled to $[0,1]$ and standardized with CLIP's channel means and standard deviations. Support masks are used later for pooling; they do not remove pixels from the encoder input.

An observation produces $P=14\times14=196$ patch features. We use column vectors throughout: $\mathbf h_p^{(\ell)}\in\R^{D_h}$ is the token after block $\ell=9$ of the 12-block transformer, with $D_h=768$. The existing final layer normalization and projection $W_{\mathrm{proj}}\in\R^{D_h\times D}$ give
\begin{equation}
 \mathbf z_p=\norm{W_{\mathrm{proj}}^\top\operatorname{LN}_{\mathrm{post}}(\mathbf h_p^{(\ell)})},
 \qquad \mathbf z_p\in\R^D,\quad D=512,
 \label{eq:token}
\end{equation}
where $\operatorname{norm}(\mathbf v)=\mathbf v/\max(\lVert\mathbf v\rVert_2,10^{-12})$. Unit normalization makes matching depend on feature direction. We write $Z=F_\ell(I)\in\R^{P\times D}$ for the resulting feature matrix, whose row $p$ is $\mathbf z_p^\top$. Patch indices follow row-major order: $p=14a+b$ for grid row $a$ and column $b$, both in $\{0,\ldots,13\}$.

The class token participates in the transformer but is omitted from $Z$. Encoding uses FP32; normalized features are cached in FP16 and restored to FP32 for pooling and scoring. Section~\ref{sec:robustness} evaluates another backbone.

\subsection{Source-balanced support prototypes}
For each support, we first resize its mask to $224\times224$ by nearest-neighbor sampling, obtaining $\widetilde M_i$. Let $\Omega_p$ be the $16\times16$ pixel region of patch $p$. Its foreground or background weight is the fraction of pixels with the corresponding label:
\begin{equation}
 w_{i,p}^{r}=\frac{1}{256}\sum_{u\in\Omega_p}\mathbf1[\widetilde M_i(u)=m_r],
 \qquad m_{\mathrm{fg}}=1,\quad m_{\mathrm{bg}}=0.
 \label{eq:coverage}
\end{equation}
Here and below, $r\in\{\mathrm{fg},\mathrm{bg}\}$. For example, a patch containing 64 foreground pixels has foreground weight $64/256=0.25$. Ignored pixels contribute to neither numerator, so the two weights can sum to less than one. Fractional weights retain partially covered patches rather than assigning each patch a single region label.

We then compute one unit vector for each region of each support image:
\begin{equation}
 \widetilde w_{i,p}^{r}=\frac{w_{i,p}^{r}}{\sum_{j=0}^{P-1}w_{i,j}^{r}},
 \qquad
 \mathbf u_i^r=\norm{\sum_{p=0}^{P-1}\widetilde w_{i,p}^{r}\mathbf z_{i,p}}.
 \label{eq:source}
\end{equation}
The normalized weights define a regional mean; the outer normalization converts that mean to a direction. Every selected support has positive foreground and background weight after resizing, so both regional means are defined. Finally, we average the $K$ support directions with equal weight and normalize again:
\begin{equation}
 \boldsymbol\mu_c^r=\norm{\frac1K\sum_{i=1}^{K}\mathbf u_i^r},
 \qquad \boldsymbol\mu_c^r\in\R^D.
 \label{eq:proto}
\end{equation}
Thus, each image contributes one foreground direction and one background direction, regardless of its region sizes. This is the source balancing used by PCLM: a source is one support image.

\subsection{Prototype contrast}
The foreground prototype describes the demonstrated category; the background prototype provides a visual reference. Their difference defines a shared scoring direction:
\begin{equation}
 \begin{aligned}
 \mathbf d_c&=\boldsymbol\mu_c^{\mathrm{fg}}-\boldsymbol\mu_c^{\mathrm{bg}},\\
 s_c(\mathbf z)&=\mathbf z^\top\mathbf d_c
 =\mathbf z^\top\boldsymbol\mu_c^{\mathrm{fg}}
  -\mathbf z^\top\boldsymbol\mu_c^{\mathrm{bg}}.
 \end{aligned}
 \label{eq:score}
\end{equation}
We retain $\mathbf d_c$ without further normalization. For unit features, the score is foreground cosine similarity minus background cosine similarity. It also equals a difference of squared distances:
\begin{equation}
 s_c(\mathbf z)=\tfrac12\left(\lVert\mathbf z-\boldsymbol\mu_c^{\mathrm{bg}}\rVert_2^2
                  -\lVert\mathbf z-\boldsymbol\mu_c^{\mathrm{fg}}\rVert_2^2\right).
 \label{eq:distance}
\end{equation}
A positive score means closer agreement with the foreground prototype, a negative score favors the background reference, and zero means equal similarity. These signed scores use the same scale in every window, allowing their relative strengths to be preserved during fusion.

\subsection{Independent local magnification}
The query is first resized to a $224\times224$ RGB canvas $C$. Canvas coordinates $u=(x,y)$ specify column $x$ and row $y$, each in $\{0,\ldots,223\}$. The fixed set $\mathcal W$ contains nine $112\times112$ windows, with upper-left coordinates $(x_W,y_W)\in\{0,56,112\}^2$. Adjacent windows overlap by 50\% along each axis. Every window is enlarged to $224\times224$ and independently encoded as $Z_W=F_\ell(C|_W)$. The main prediction fuses these nine local views; the full-canvas view is evaluated separately as a comparison.

A global observation has a nominal patch footprint of 16 canvas pixels per side. Enlarging a half-image window reduces that footprint to eight pixels per side, so a visible target occupies approximately four times as many patch areas. Overlap also presents a location in several crop contexts. All windows come from the same initial canvas: magnification changes the encoder's sampling and context while keeping the available source image information fixed.

\subsection{Reprojection and coverage fusion}
The feature matrix of window $W$ first produces a raw $14\times14$ score map $R_W$. We smooth this map before placing it back on the canvas:
\begin{equation}
 R_W[a,b]=\mathbf z_{W,14a+b}^{\top}\mathbf d_c,
 \qquad \widehat R_W=\mathcal M_3(R_W),
 \label{eq:patchmaps}
\end{equation}
where $\mathcal M_3$ is a $3\times3$ mean filter with replicated boundary values. It averages neighboring patch scores within each view. The score scale is retained throughout this operation.

Nearest-neighbor reprojection expands each filtered patch score to an $8\times8$ region in its original window. Define $S_W$ directly in canvas coordinates, for pixels covered by $W$:
\begin{equation}
 S_W(x,y)=\widehat R_W\!\left[
       \left\lfloor\frac{y-y_W}{8}\right\rfloor,
       \left\lfloor\frac{x-x_W}{8}\right\rfloor\right],
 \qquad (x,y)\in W.
 \label{eq:reprojection}
\end{equation}
The first array index is a row and the second a column. This single step both expands the map to $112\times112$ and places it at the window's canvas location.

Let $\mathcal W(u)=\{W\in\mathcal W:u\in W\}$ be the windows covering canvas pixel $u$. The final heatmap is their mean:
\begin{equation}
 H_c(u)=\frac{1}{|\mathcal W(u)|}\sum_{W\in\mathcal W(u)}S_W(u).
 \label{eq:aggregation}
\end{equation}
Every pixel is covered by one, two or four windows. Dividing by this local count gives all pixels the same score scale, even though their numbers of observations differ. The output remains constant within aligned $8\times8$ canvas cells.

For fixed features and windows, scoring, mean filtering, reprojection and coverage averaging are linear. Equivalently, $H_c(u)=\mathbf d_c^\top\overline{\mathbf z}_q(u)$, where $\overline{\mathbf z}_q(u)$ is the corresponding weighted average of local features, without another normalization. The class direction therefore acts consistently on the combined observations. Section~\ref{sec:controls} tests alternative score transformations before fusion.

\subsection{Algorithm and computational cost}
Algorithm~\ref{alg:pclm} summarizes the two stages. After encoding, support construction costs $O(KPD)$, with $P=196$ and $D=512$. Storing only the class direction requires $D$ FP32 values, or 2 KiB. The implementation used for timing retains both prototype centers, using 4 KiB per class. Either representation has fixed size as the support count grows.

For $V=9$ windows, query scoring costs $O(VPD)$; fusion is linear in the total reprojected map size. The independent windows are encoded in one batch, and encoder computation dominates latency. Adding supports changes the estimated prototypes while leaving query-time storage and dot-product work unchanged. Adding windows increases query computation while reusing the same class direction. Section~\ref{sec:local} compares this observation strategy with higher-resolution global encoding.

\begin{algorithm}[tbp]
\caption{PCLM: class construction and query localization}
\label{alg:pclm}
\begin{spacing}{1}\normalsize
\begin{algorithmic}[1]
\Require Frozen feature routine $F_\ell$; support set $\mathcal S_c$; query $I_q$
\Ensure Reusable direction $\mathbf d_c\in\R^{512}$ and heatmap $H_c\in\R^{224\times224}$
\Statex \textit{Class construction (once per category)}
\For{each support $(I_i,M_i)\in\mathcal S_c$}
 \State $Z_i\gets F_\ell(I_i)$; $\widetilde M_i\gets\operatorname{Resize}_{224}^{\mathrm{nearest}}(M_i)$
 \State Compute $w_{i,p}^r$ and $\mathbf u_i^r$ for both regions using Eqs.~\eqref{eq:coverage}--\eqref{eq:source}
\EndFor
\State $\boldsymbol\mu_c^r\gets\norm{K^{-1}\sum_i\mathbf u_i^r}$ for $r\in\{\mathrm{fg},\mathrm{bg}\}$
\State Store $\mathbf d_c\gets\boldsymbol\mu_c^{\mathrm{fg}}-\boldsymbol\mu_c^{\mathrm{bg}}$
\Statex \textit{Query localization}
\State $C\gets\operatorname{Resize}_{224}^{\mathrm{bicubic}}(I_q)$; initialize canvas arrays $A\gets0$, $B\gets0$
\For{each of the nine windows $W\in\mathcal W$}
 \State $Z_W\gets F_\ell(C|_W)$ \Comment{enlarge and independently encode}
 \State $R_W\gets\operatorname{reshape}_{14\times14}(Z_W\mathbf d_c)$
 \State $\widehat R_W\gets\mathcal M_3(R_W)$ \Comment{$3\times3$ mean, replicated boundaries}
 \State Place $S_W$ on its canvas window using Eq.~\eqref{eq:reprojection}
 \State $A(u)\gets A(u)+S_W(u)$; $B(u)\gets B(u)+1$ for every $u\in W$
\EndFor
\State $H_c(u)\gets A(u)/B(u)$ for every canvas pixel $u$
\State \Return $\mathbf d_c,H_c$
\end{algorithmic}\end{spacing}
\end{algorithm}
\FloatBarrier

\section{Experimental protocol}
\label{sec:protocol}
\subsection{Queries, supports and category transfer}
A query is a class--image pair. An image containing several evaluated categories contributes one query and one heatmap for each category. Table~\ref{tab:data} lists the evaluation sets: VOC 2012 validation~\cite{everingham2010voc,everingham2015retrospective}, COCO 2017 validation~\cite{lin2014coco}, ADE20K validation~\cite{zhou2017ade20k,zhou2019semantic} and Oxford-IIIT Pets test~\cite{parkhi2012pets}. We refer to VOC, COCO and ADE20K as the scene datasets. Pets breeds are mapped to cat or dog, making its task foreground localization at the category level.

All support images and masks come from VOC training. We construct five support banks with seeds 11, 23, 37, 53 and 71. Each prediction uses five support images per class from one bank; the five predictions are evaluated separately and their metrics are averaged. Support selection excludes all development and validation images. For the support-budget study, the first 1, 2, 5, 10 or 20 eligible images in each saved candidate sequence form nested support sets. Eligibility requires both known foreground and known background after resizing.

\begin{table}[t]
\centering
\caption{Evaluation queries after fixed label conversion and eligibility checks. Small denotes category-union foreground below 5\% of the full $224\times224$ canvas. All supports are from VOC training.}
\label{tab:data}
\begingroup
\renewcommand{\baselinestretch}{1}\normalsize
\renewcommand{\arraystretch}{1.12}
\setlength{\tabcolsep}{3pt}

\begin{tabular*}{\linewidth}{@{\extracolsep{\fill}}>{\raggedright\arraybackslash}p{.32\linewidth}rrrr@{}}
\toprule Dataset & Images & Query pairs & Categories & Small pairs\\\midrule
VOC validation & 1,449 & 2,147 & 20 & 655\\
COCO validation & 4,031 & 6,900 & 20 & 2,978\\
ADE20K validation & 1,218 & 2,015 & 12 & 1,338\\
Pets test & 3,662 & 3,662 & 2 & 76\\
\bottomrule\end{tabular*}
\endgroup
\end{table}

The category mapping retains the 20 VOC categories in COCO, 12 mapped categories in ADE20K, and cat and dog in Pets. A target mask is the union of all instances of the requested category. VOC void pixels and ADE20K label 0 are ignored; Pets trimap labels 1/2/3 map to foreground/background/ignore. COCO unions include positive-area crowd annotations of the target category. We exclude nine query examples: seven Pets images with background-only trimaps and two scene queries whose foreground vanishes after resizing. Query annotations are used to compute evaluation metrics and target-area groups; inference uses the RGB image and the stored class direction.

\subsection{Development and implementation settings}
We use an additional 220 labeled VOC training class--image pairs to select the method settings. The first 60 pairs compare 192 combinations of feature layer, prototype scoring and patch filtering. A separate set of 160 images compares window sizes and fusion rules. These development masks measure the accuracy of candidate settings; the five support images and masks per class construct the prototypes used by each candidate. Development images are excluded from support selection. Window selection maximizes small-target AP while retaining at least 95\% of global overall AP, averaged over support banks 11 and 23. The selected settings are held fixed for benchmark evaluation.

The main configuration uses OpenAI CLIP ViT-B/16, block 9, five supports per class, nine $112\times112$ windows and a $3\times3$ mean filter on each patch map. Each window is enlarged to $224\times224$. RGB images use bicubic resizing and masks use nearest-neighbor resizing. Encoding uses FP32; normalized features are stored in FP16 and restored to FP32 for pooling and scoring. Comparisons use the same checkpoint, queries and category mappings. Text-conditioned baselines retain their own spatial scoring operators, and PCLM scoring ablations share support images, masks and frozen features. CCI uses seven final-block clusters with ten K-means initializations and seed 20260907. CCI and Grad-ECLIP use a single photo template, while CLIP Surgery and Grad-CAM use an 85-template ensemble. LeGrad uses the last two blocks and its native normalization; gScoreCAM uses last-block activations, CLS-gradient channel selection, $k=300$ and attribution batch 64 with native score weighting. Grad-ECLIP retains raw nonnegative relevance. These settings are fixed independently of query accuracy.

\subsection{Metrics and statistical analysis}
We compute pixel average precision (AP) separately for each query over its known foreground and background pixels, grouping equal scores at the same precision--recall threshold. Dataset AP is the mean over queries. A small target has a class-union foreground area below 5\% of the full resized canvas. This image-level category area differs from COCO's instance-area definition. We also report the disjoint area bins $<1\%$, $[1\%,2\%)$ and $[2\%,5\%)$.

Pointing measures whether a highest-scoring pixel lies in the foreground. We report first-maximum Pointing, $P_1$, which selects the first maximum among known pixels in row-major order. All metrics use scores before display color scaling.

For PCLM and its ablations, we average each query's metrics across the five banks before computing dataset means. Paired 95\% confidence intervals use 10,000 bootstrap samples of original images~\cite{efron1994bootstrap}; all category queries from an image remain together in each sample. We also summarize the range of paired differences across banks to measure sensitivity to support selection. Means and gains are rounded separately from the original records, so subtracting printed means can differ from a reported gain by 0.01 percentage points.

\subsection{Timing and reproducibility}
We measure latency on an RTX 4090 with an AMD EPYC 7642 host, using ten warmups and 50 fixed VOC images. CUDA is synchronized before and after each measurement. RGB-to-heatmap time includes in-memory preprocessing, encoding, feature transfer, scoring and CPU output. Disk decoding and one-time support construction are excluded. Localization baselines, view layouts and resolution selection have separate timing series.

% Results connect comparative accuracy, local observation, contrast and reuse.
\section{Results and analysis}
We first compare PCLM with text-conditioned localization methods and inspect their heatmaps. We then hold the support representation fixed to study local observation, and hold the observations fixed to study prototype contrast. The final experiments examine category reuse, support and view budgets, and robustness to encoder and mask changes.

\subsection{Comparative small-target localization}
\label{sec:comparisons}
Table~\ref{tab:text} compares PCLM with CLIP localization baselines on all 5,047 small-target queries. The baselines specify the category with class text, while PCLM uses five masked VOC supports per class. PCLM achieves the highest small-target AP in this comparison on every dataset: 63.99 on VOC, 41.71 on COCO, 36.86 on ADE20K and 97.01 on Pets.

\begin{table}[t]
\centering
\caption{Small-target AP (\%) for text-conditioned localization adapters and PCLM. The text block receives class names; PCLM receives five masked supports. ``+ nine'' denotes the nine-window extension.}
\label{tab:text}
\begingroup
\renewcommand{\baselinestretch}{1}\normalsize
\renewcommand{\arraystretch}{1.12}
\setlength{\tabcolsep}{3pt}

\begin{tabular*}{\linewidth}{@{\extracolsep{\fill}}>{\raggedright\arraybackslash}p{.48\linewidth}rrrr@{}}
\toprule Approach and observation & VOC & COCO & ADE20K & Pets\\\midrule
CCI, global & 11.57 & 5.61 & 3.71 & 47.67\\
Grad-CAM, global & 23.84 & 18.90 & 19.13 & 13.50\\
Grad-ECLIP, global & 35.88 & 22.46 & 20.88 & 73.05\\
Grad-ECLIP + nine & 45.84 & 29.12 & 26.96 & 86.26\\
CLIP Surgery, global & 37.06 & 26.01 & 22.41 & 63.20\\
CLIP Surgery + nine & 50.76 & 36.08 & 30.63 & 81.72\\
LeGrad, global & 43.00 & 27.77 & 24.46 & 81.83\\
LeGrad + nine & 34.17 & 19.58 & 17.75 & 84.39\\
gScoreCAM, global & 8.24 & 4.86 & 3.37 & 5.49\\
\midrule
PCLM, nine views & 63.99 & 41.71 & 36.86 & 97.01\\
\bottomrule\end{tabular*}
\endgroup
\end{table}

Nine-view CLIP Surgery is the strongest baseline on the three scene datasets. PCLM improves on it by 13.23, 5.63 and 6.23 AP points on VOC, COCO and ADE20K, respectively. Both methods use the same nine-window layout. On Pets, PCLM exceeds the strongest baseline, nine-view Grad-ECLIP, by 10.75 points.

Adding local views has different effects on the baselines: nine windows improve Grad-ECLIP on the scene datasets but reduce LeGrad AP. The scoring rule therefore matters when combining enlarged views. Section~\ref{sec:controls} tests PCLM's shared foreground--background score.

\subsection{Qualitative analysis}
\label{sec:qualitative}
\begin{figure}[!tb]
\centering\includegraphics[width=\linewidth]{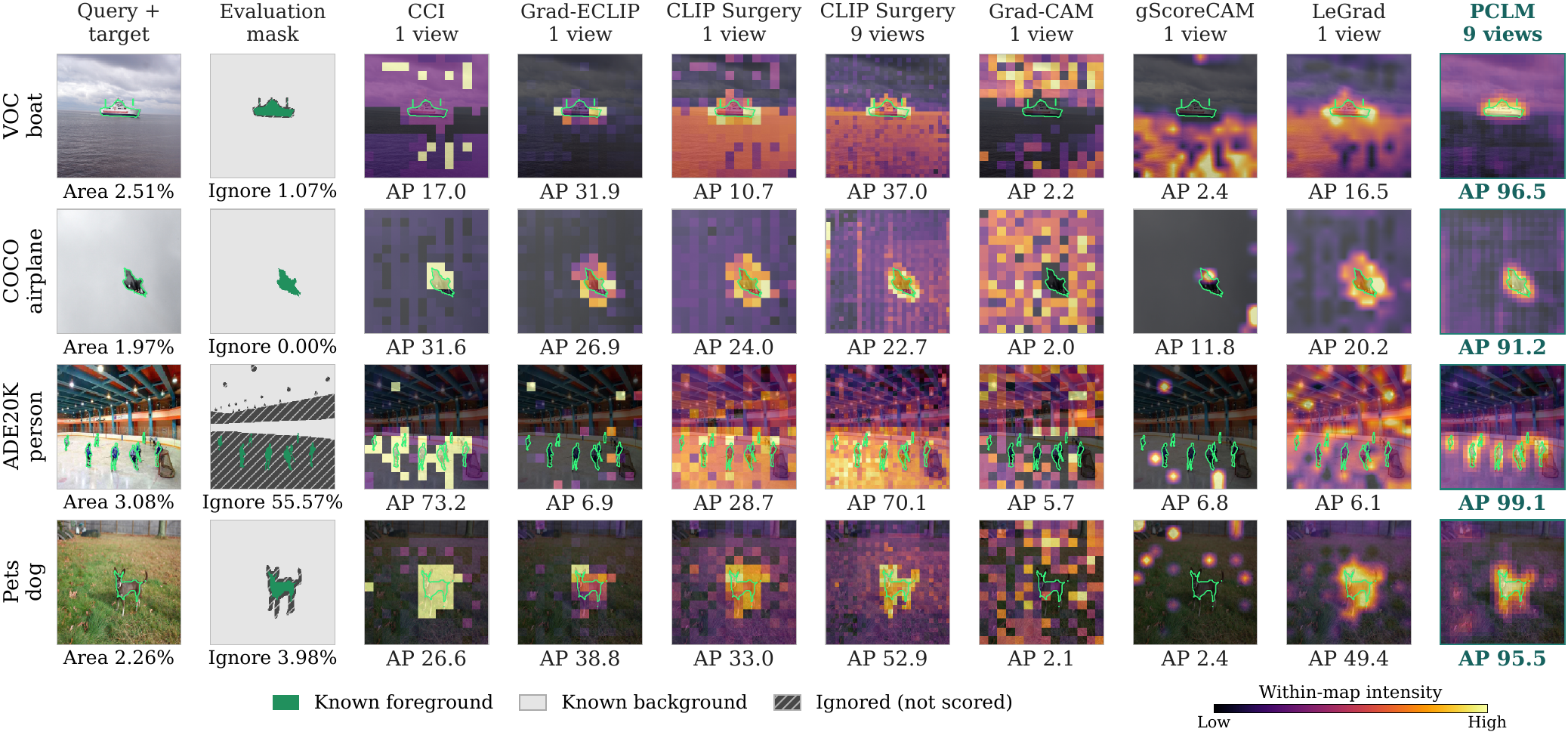}
\caption{Selected successful small-target localizations across four datasets. PCLM uses five supports from bank 11 and nine views; baseline view counts are shown above each column. CCI uses final-block masking and signed cluster-score normalization. Green denotes foreground, light gray background and hatched gray ignored pixels. Per-map 1st--99th percentile scaling affects display only; AP uses scores before display scaling.}
\label{fig:success}
\end{figure}

Figure~\ref{fig:success} compares successful localizations across all four datasets. PCLM concentrates responses on small objects and distinguishes spatially separated instances of the same category. The evaluation masks identify the foreground, background and ignored pixels used to score each example.

\begin{figure}[!tb]
\centering\includegraphics[width=\linewidth]{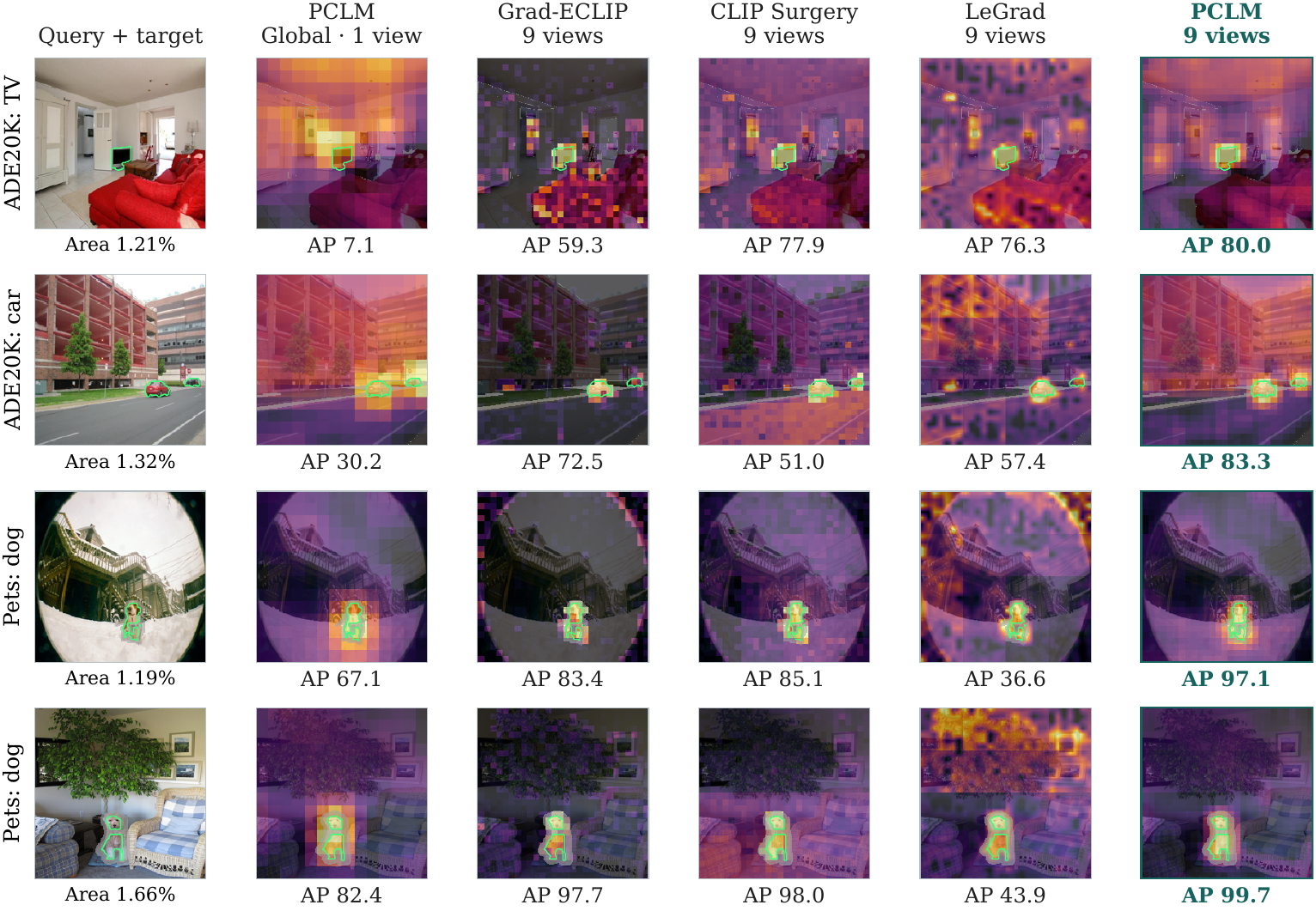}
\caption{Selected successful ADE20K and Pets examples. PCLM uses five masked supports from bank 11; the references use class text. All local maps use nine windows, retaining each method's per-view scores before projection and coverage averaging. Green contours mark targets and gray shading marks ignored pixels. Display uses per-map 1st--99th percentile scaling; AP uses unscaled scores over known pixels.}
\label{fig:cases}
\end{figure}

Figure~\ref{fig:cases} compares PCLM with Grad-ECLIP, CLIP Surgery and LeGrad under the same nine-window layout, alongside global PCLM with the same support bank. In ADE20K, local PCLM concentrates the response on a television occupying 1.21\% of the canvas and resolves two separated cars from the broader global response. For the cars, PCLM reaches 83.3 AP, compared with 72.5 for Grad-ECLIP, 51.0 for CLIP Surgery and 57.4 for LeGrad.

The Pets examples show concentrated responses around the dog in both outdoor and indoor scenes, reaching 97.1 and 99.7 AP. Grad-ECLIP and CLIP Surgery also clearly localize the indoor dog, while LeGrad additionally highlights foliage and wall regions. These selected examples show successful localization of visible targets; the additional panel also requires an improvement over global PCLM.

\subsection{Local magnification versus whole-image enlargement}
\label{sec:local}
\label{sec:overall}
We keep the support prototypes, feature layer and scoring direction fixed and change only how CLIP observes the query. The comparisons cover global 224, four corner windows and nine overlapping windows, followed by controls for resolution, source-image detail and filtering.

\paragraph{Localization gains and target scale}
Table~\ref{tab:main} shows that nine windows improve scene-dataset overall AP by 8.09--9.93 points and small-target AP by 12.52--19.73 points. Four windows already provide large improvements; the center and edge-centered windows add a further 3.92, 2.64 and 1.93 small-target AP points on VOC, COCO and ADE20K. The larger gains on small targets are consistent with their limited patch coverage in the global view.

\begin{table}[t]
\centering
\caption{PCLM localization with fixed five-shot supports. AP is percent. G/4/9 denote global 224, four corners and nine windows. Gain and paired 95\% interval are nine minus global in percentage points. Means average five banks.}
\label{tab:main}
\begingroup
\renewcommand{\baselinestretch}{1}\normalsize
\renewcommand{\arraystretch}{1.12}
\setlength{\tabcolsep}{3pt}

\begin{tabular*}{\linewidth}{@{\extracolsep{\fill}}>{\raggedright\arraybackslash}p{.25\linewidth}rrrrr@{}}
\toprule Query set & G & 4 & 9 & Gain & 95\% CI\\\midrule
VOC, all & 73.14 & 78.76 & 81.35 & 8.21 & $[7.61,8.82]$\\
VOC, small & 44.25 & 60.07 & 63.99 & 19.73 & $[18.38,21.15]$\\
COCO, all & 54.51 & 60.15 & 62.60 & 8.09 & $[7.77,8.40]$\\
COCO, small & 27.70 & 39.07 & 41.71 & 14.00 & $[13.45,14.56]$\\
ADE20K, all & 38.78 & 46.84 & 48.71 & 9.93 & $[9.30,10.57]$\\
ADE20K, small & 24.34 & 34.93 & 36.86 & 12.52 & $[11.71,13.34]$\\
Pets, all & 97.91 & 98.86 & 99.27 & 1.36 & $[1.28,1.44]$\\
Pets, small & 88.58 & 95.14 & 97.01 & 8.44 & ---\\
\bottomrule\end{tabular*}
\endgroup
\end{table}

\begin{figure}[!tb]
\centering\includegraphics[width=\linewidth]{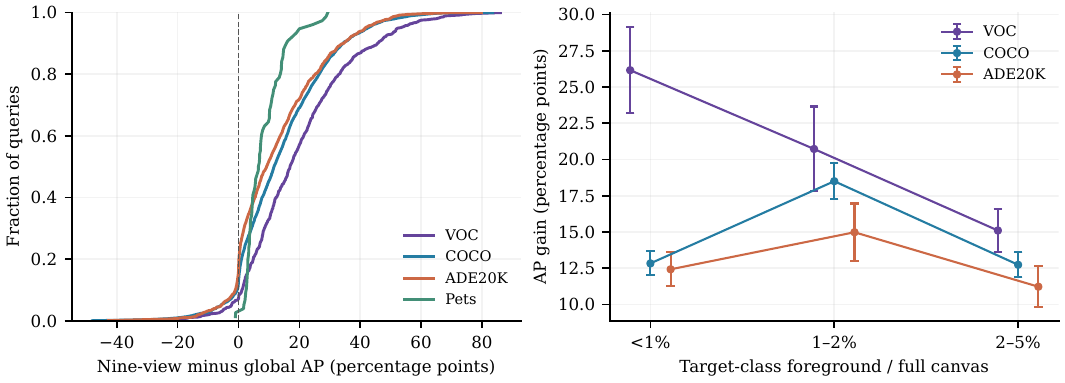}
\caption{Who benefits from local magnification? Left: empirical cumulative distributions of nine-minus-global AP changes after averaging five support banks. Right: scene-dataset mean gains by target-area bin with paired parent-image 95\% intervals. Area bins are $<1\%$, $[1\%,2\%)$ and $[2\%,5\%)$ of the canvas.}
\label{fig:gain}
\end{figure}

Figure~\ref{fig:gain} shows how these gains vary across queries and target sizes. AP improves for 92.82\%, 87.51\% and 83.86\% of small queries on the three scene datasets, with median gains of 16.96, 11.69 and 9.20 points. For targets below 1\% of the canvas, AP rises from 19.74 to 45.90 on VOC, 9.91 to 22.74 on COCO and 9.63 to 22.05 on ADE20K. Gains remain positive in both larger area bins. The improvement is shared by most small-target queries.

\paragraph{Observation geometry with matched filtering}
A $3\times3$ patch filter covers a square about 48 canvas pixels wide in the global view and 24 pixels wide in an enlarged local view. To match smoothing scales, we replace the per-view patch filter with a $25\times25$ filter applied after projection and fusion. Table~\ref{tab:local} shows that local gains of 7.09, 4.59 and 4.22 points remain on the three scene datasets. Local re-encoding therefore improves accuracy even when global and local outputs receive the same spatial smoothing.

\begin{table}[t]
\centering
\caption{View layouts and common-coordinate filtering with fixed PCLM prototypes: small-target AP (\%). Patch 3 applies a $3\times3$ filter on each patch map; Canvas 25 applies a $25\times25$ filter in output coordinates.}
\label{tab:local}
\begingroup\tablesize
\begin{tabular*}{\linewidth}{@{\extracolsep{\fill}}lrrrr@{}}
\toprule Observation and filter & VOC & COCO & ADE20K & Pets\\\midrule
Global 224, patch 3 & 44.25 & 27.70 & 24.34 & 88.58\\
Four windows, patch 3 & 60.07 & 39.07 & 34.93 & 95.14\\
Nine windows, patch 3 & 63.99 & 41.71 & 36.86 & 97.01\\
\midrule
Global 224, canvas 25 & 59.30 & 38.92 & 34.12 & 94.88\\
Nine windows, canvas 25 & 66.39 & 43.52 & 38.34 & 97.73\\
\bottomrule\end{tabular*}
\endgroup
\end{table}

\paragraph{Whole-image resolution and comparable computation}
Table~\ref{tab:resolution} compares local observation with two global input paths. \emph{Canvas} inputs enlarge the same $224\times224$ image used by PCLM. \emph{Original RGB} inputs resize the source image directly, preserving detail that the intermediate canvas may remove. The checkpoint, feature layer, support prototypes and output coordinates are shared; spatial positional embeddings are interpolated bicubically.

\begin{table}[tbp]
\def\baselinestretch{1}\selectfont
\centering\normalsize
\setlength{\tabcolsep}{1.5pt}
\caption{Global resolution, input source and spatial filtering. (a) Small-target AP / first-maximum Pointing (\%). (b) Nine-view minus global paired AP gain and 95\% interval under patch-3 filtering, in percentage points. All five banks and all 5,047 queries are used. Canvas inputs enlarge the shared 224-pixel image; Original RGB inputs resize the source image directly. Global resolution is an input side length; 9 views use nine independently encoded $224\times224$ forwards.}
\label{tab:resolution}
\begin{tabular*}{\linewidth}{@{\extracolsep{\fill}}lllrrrr@{}}
\toprule \multicolumn{7}{l}{\textbf{(a) Resolution, image source and filtering}}\\
Input & Source & Filter & VOC & COCO & ADE20K & Pets\\\midrule
448 & Canvas & None & 56.47 / 62.26 & 36.44 / 38.19 & 32.71 / 31.91 & 88.41 / 91.05\\
448 & Canvas & Patch 3 & 59.89 / 71.42 & 37.57 / 45.94 & 33.81 / 40.01 & 96.37 / 99.74\\
448 & Canvas & Canvas 25 & 61.51 / 71.48 & 38.66 / 46.82 & 34.79 / 40.22 & 97.21 / 100.00\\
448 & Original & None & 57.26 / 64.03 & 37.91 / 40.20 & 33.69 / 33.24 & 88.36 / 93.42\\
448 & Original & Patch 3 & 61.09 / 72.82 & 39.34 / 47.90 & 35.08 / 41.49 & 96.87 / 100.00\\
448 & Original & Canvas 25 & 62.93 / 72.98 & 40.66 / 48.32 & 36.19 / 41.67 & 97.78 / 100.00\\
640 & Canvas & None & 54.47 / 64.15 & 34.15 / 40.46 & 30.04 / 34.75 & 92.77 / 94.47\\
640 & Canvas & Patch 3 & 58.30 / 69.07 & 36.77 / 46.87 & 32.35 / 39.84 & 97.68 / 97.63\\
640 & Canvas & Canvas 25 & 54.49 / 64.70 & 33.60 / 40.91 & 29.99 / 34.57 & 97.52 / 100.00\\
640 & Original & None & 57.74 / 67.94 & 37.71 / 43.88 & 32.61 / 35.92 & 93.37 / 98.16\\
640 & Original & Patch 3 & 62.27 / 74.02 & 40.84 / 51.40 & 35.28 / 42.39 & 98.36 / 100.00\\
640 & Original & Canvas 25 & 58.51 / 68.98 & 37.31 / 44.62 & 32.54 / 37.56 & 98.19 / 100.00\\
\midrule
9 views & Canvas & None & 59.83 / 65.71 & 40.14 / 42.04 & 35.59 / 35.64 & 90.64 / 93.16\\
9 views & Canvas & Patch 3 & 63.99 / 75.91 & 41.71 / 49.54 & 36.86 / 43.00 & 97.01 / 99.47\\
9 views & Canvas & Canvas 25 & 66.39 / 77.16 & 43.52 / 50.64 & 38.34 / 44.17 & 97.73 / 99.21\\
\bottomrule\end{tabular*}\par\medskip
\textbf{(b) Nine-view \method minus global \method: paired AP gain [95\% CI], pp}\par\smallskip
\label{tab:resolution-contrasts}
\begin{tabular*}{\linewidth}{@{\extracolsep{\fill}}lrrrr@{}}
\toprule \shortstack{Global\\comparator} & VOC & COCO & ADE20K & Pets\\\midrule
448 / canvas & 4.10 $[3.14,5.09]$ & 4.13 $[3.72,4.56]$ & 3.05 $[2.47,3.62]$ & 0.64 $[-0.47,2.01]$\\
448 / native & 2.90 $[1.98,3.83]$ & 2.37 $[1.97,2.78]$ & 1.78 $[1.22,2.34]$ & 0.14 $[-0.73,0.90]$\\
640 / canvas & 5.68 $[4.26,7.11]$ & 4.94 $[4.34,5.57]$ & 4.51 $[3.69,5.36]$ & -0.66 $[-2.05,0.84]$\\
640 / native & 1.71 $[0.46,2.98]$ & 0.87 $[0.33,1.41]$ & 1.58 $[0.79,2.38]$ & -1.34 $[-2.47,-0.45]$\\
\bottomrule\end{tabular*}
\end{table}

At $448\times448$, global inference matches the nominal eight-pixel canvas patch footprint of half-image magnification. Nine windows exceed canvas-448 by 4.10, 4.13 and 3.05 scene AP points. At comparable measured latency, the gains over canvas-640 are 5.68, 4.94 and 4.51 points. We select input side length 640 because its mean latency is closest to that of nine views among $\{320,384,448,512,576,640,672\}$, using timing alone. It takes 28.71 ms versus 27.67 ms for nine views on the same fifty images. Local re-encoding thus extracts more useful small-target evidence from the available canvas than whole-image enlargement at similar cost.

Direct access to original RGB further strengthens global inference. Resizing original RGB directly to 640 (native-640) reaches 62.27, 40.84 and 35.28 AP on the scene datasets, versus 63.99, 41.71 and 36.86 for nine views; Pets favors native-640, at 98.36 versus 97.01. The corresponding scene-dataset paired intervals remain positive, as shown in Table~\ref{tab:resolution}. A source-input timing check gives 27.61 ms for native-640 and 27.47 ms for nine views. Both access to source detail and the choice of observation layout affect accuracy. Local views retain a positive gain on the scene datasets, while the higher-resolution original-image input performs better on Pets.

\subsection{Prototype contrast and multiview scoring}
\label{sec:controls}
To measure the effect of background subtraction, we compare the full score with foreground similarity alone. Both use the same PCLM foreground prototype, support images, features, views and filter. Table~\ref{tab:controls} shows that nine-view foreground--background scoring exceeds foreground-only matching by 27.79, 22.21 and 20.73 AP points on the scene datasets. Subtracting background similarity improves the ranking of target pixels within the enlarged views.

\begin{table}[t]
\centering
\caption{Main-method controls: nine-view small-target AP (\%). Foreground-only uses the exact PCLM foreground center. All score transforms are applied before patch-3 smoothing.}
\label{tab:controls}
\begingroup
\renewcommand{\baselinestretch}{1}\normalsize
\renewcommand{\arraystretch}{1.12}
\setlength{\tabcolsep}{3pt}

\begin{tabular*}{\linewidth}{@{\extracolsep{\fill}}>{\raggedright\arraybackslash}p{.48\linewidth}rrrr@{}}
\toprule Scoring / fusion & VOC & COCO & ADE20K & Pets\\\midrule
Foreground similarity only & 36.20 & 19.49 & 16.13 & 84.90\\
Foreground--background raw margin & 63.99 & 41.71 & 36.86 & 97.01\\\midrule
Per-view min--max margin & 46.16 & 27.10 & 24.99 & 91.18\\
Per-view standardized margin & 56.73 & 36.72 & 32.76 & 92.27\\
$\operatorname{sigmoid}(20s)-0.5$ & 64.03 & 41.87 & 37.05 & 97.11\\
\bottomrule\end{tabular*}
\endgroup
\end{table}

\paragraph{Interaction between contrast and local observation}
We combine two scores (foreground-only and foreground--background) with three observation layouts (global, four and nine views) and three filters. Table~\ref{tab:interaction} measures how much contrast changes the gain from adding local views: it subtracts the foreground-only global-to-nine gain from the contrastive global-to-nine gain. This interaction is positive on every scene dataset under each filter. Even with the common canvas filter, contrast adds 3.45, 2.96 and 3.35 points to the gain from local observation. Without smoothing, foreground-only nine-view AP decreases relative to global inference, while PCLM improves. Thus, local re-encoding is more effective when patches are scored relative to a background reference.

% Source: app__table_10.tex, scene-dataset interaction estimates and paired CIs.
% Means independently checked against archived small_mechanism_arrays.npz.
\begin{table}[t]
\centering
\caption{Interaction between foreground--background contrast and local observation on small scene targets (AP percentage points). The interaction is $(P_9-P_G)-(F_9-F_G)$, where $P$ is PCLM and $F$ uses its foreground center alone. Each cell gives the interaction and paired parent-image 95\% interval. Canvas 25 applies the same image-coordinate filter to global and local maps.}
\label{tab:interaction}
\begingroup
\renewcommand{\baselinestretch}{1}\normalsize
\renewcommand{\arraystretch}{1.15}
\setlength{\tabcolsep}{3pt}
\begin{tabular*}{\linewidth}{@{\extracolsep{\fill}}lccc@{}}
\toprule
Dataset & No smoothing & Patch $3\times3$ & Canvas $25\times25$\\
\midrule
VOC & \shortstack{17.58\\$[16.54,18.65]$} & \shortstack{15.02\\$[13.64,16.45]$} & \shortstack{3.45\\$[1.99,4.89]$}\\
\addlinespace[3pt]
COCO & \shortstack{12.36\\$[11.94,12.79]$} & \shortstack{11.91\\$[11.37,12.45]$} & \shortstack{2.96\\$[2.44,3.49]$}\\
\addlinespace[3pt]
ADE20K & \shortstack{10.85\\$[10.18,11.51]$} & \shortstack{11.52\\$[10.73,12.31]$} & \shortstack{3.35\\$[2.58,4.11]$}\\
\bottomrule
\end{tabular*}
\endgroup
\end{table}

\paragraph{Keeping a common score scale across windows}
PCLM applies the same support-derived direction to every view and averages the scores at each covered location. We test this choice by applying per-window min--max, standardization or a fixed sigmoid before filtering and fusion. Relative to raw margins, min--max lowers scene AP by 11.87--17.83 points and standardization by 4.10--7.25 points (Table~\ref{tab:controls}). The fixed sigmoid stays close to the raw margin. Per-window rescaling can give a weak response the same numerical range as a strong response. Averaging the original margins retains their relative magnitudes.

\subsection{Category reuse with different support and view budgets}
\label{sec:shots}
PCLM separates category construction from query observation. Five masked VOC training images per class construct a direction that is reused for VOC validation and transferred unchanged to mapped COCO, ADE20K and Pets categories. The stored prototype values are identical across these evaluations. Tables~\ref{tab:text} and~\ref{tab:main} therefore measure localization using the same category representation on each dataset.

\begin{figure}[!tb]
\centering\includegraphics[width=\linewidth]{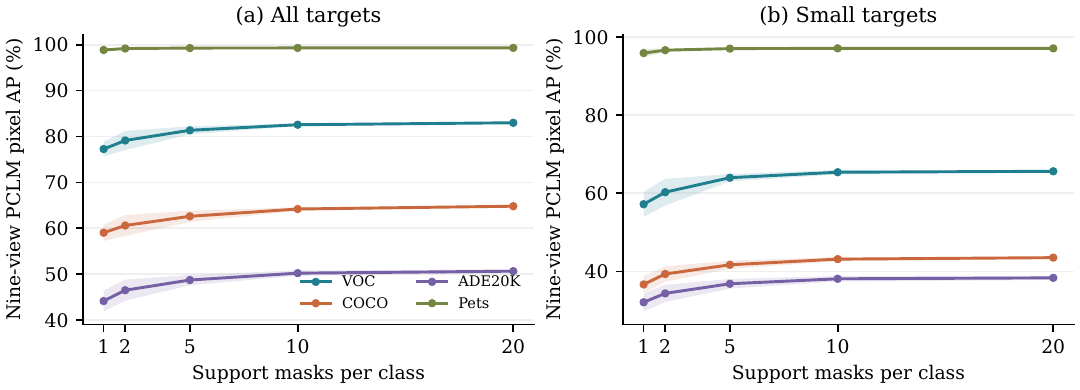}
\caption{Nine-view localization over nested support budgets. Left: all-query AP; right: small-target AP. Shading shows one sample standard deviation over five support banks. The stored class representation has the same size at every support budget.}
\label{fig:shots}
\end{figure}

\paragraph{Annotation budget}
For each bank, the 1/2/5/10/20-shot study takes the first 1, 2, 5, 10 or 20 images from the same ordered support pool. Query images and inference settings stay fixed. Figure~\ref{fig:shots} shows that more supports improve scene-dataset AP and generally reduce bank variation while class-state size stays constant. Small-target VOC AP increases from $57.20\pm3.20$ at one shot to $65.63\pm0.13$ at twenty. For overall scene AP, five-to-ten supports add 1.21--1.59 points, followed by another 0.43--0.61 from ten to twenty. The five-shot setting captures much of the benefit of the larger support sets.

\paragraph{Observation budget}
Table~\ref{tab:budget} reports the accuracy--latency trade-off for fixed window layouts. Four and nine windows take 14.78 and 27.36 ms in this series; intermediate layouts show the accuracy gained by adding center and edge views. A separate localization-baseline timing series records $26.38\pm1.31$ ms for nine-view PCLM. PCLM batches its windows and reuses the same class state across all observations.

\begin{table}[tbp]
\def\baselinestretch{1}\selectfont
\centering\normalsize
\caption{Fixed \method view layouts, complete-query AP and measured inference time. Each dataset cell is overall / small-target AP (\%). All crop layouts retain the four corners. Time covers in-memory RGB to CPU heatmap on RTX 4090, excluding support preparation and disk decoding; every row belongs to the same timing series.}
\label{tab:budget}
\setlength{\tabcolsep}{1.5pt}
\begin{tabular*}{\linewidth}{@{\extracolsep{\fill}}lrrrrrr@{}}
\toprule Layout & Views & ms & VOC & COCO & ADE20K & Pets\\\midrule
Global & 1 & 7.27 & 73.14 / 44.25 & 54.51 / 27.70 & 38.78 / 24.34 & 97.91 / 88.58\\
Four corners & 4 & 14.78 & 78.76 / 60.07 & 60.15 / 39.07 & 46.84 / 34.93 & 98.86 / 95.14\\
Corners + center & 5 & 16.74 & 80.30 / 62.58 & 61.50 / 40.59 & 47.78 / 35.90 & 99.08 / 96.59\\
\shortstack[l]{Corners +\\top/bottom} & 6 & 19.75 & 79.96 / 62.11 & 61.24 / 40.22 & 47.51 / 35.57 & 99.12 / 96.36\\
\shortstack[l]{Corners +\\left/right} & 6 & 19.94 & 80.29 / 62.28 & 61.54 / 40.65 & 48.00 / 36.15 & 99.10 / 96.25\\
Nine windows & 9 & 27.36 & 81.35 / 63.99 & 62.60 / 41.71 & 48.71 / 36.86 & 99.27 / 97.01\\

\bottomrule\end{tabular*}
\end{table}

Annotation and observation provide complementary gains. Twenty-shot global VOC AP is 74.26, while one-shot nine-view AP reaches 77.26 on the same complete query set. In this comparison, one support with nine views is more accurate than twenty supports with a global view. At five supports per class, the nine-view advantage over native-640 on scene small targets is also positive in all five support banks: paired ranges are 1.377--2.366, 0.665--1.230 and 0.968--1.967 points on VOC, COCO and ADE20K. The same category representation can therefore be used with different observation budgets, while additional supports improve its estimate.

\subsection{Backbone transfer and support-boundary robustness}
\label{sec:robustness}
We apply PCLM to CLIP ViT-L/14 at block 18, keeping the same relative feature depth and VOC support images. The prototypes are recomputed with the larger encoder. Table~\ref{tab:backbone} reports every valid VOC and ADE20K query. Nine views raise small-target AP from 49.76 to 68.82 on VOC and from 27.92 to 39.90 on ADE20K; overall AP improves by 7.04 and 8.75 points. Local observation also improves localization with the larger backbone.

\begin{table}[tbp]
\def\baselinestretch{1}\selectfont
\centering\normalsize
\caption{\method on ViT-L/14 block 18 with unchanged VOC support identities. AP is percent; gain and interval are nine-window minus global overall AP in percentage points. Every valid VOC and ADE20K query is evaluated.}
\label{tab:backbone}
\setlength{\tabcolsep}{2pt}
\begin{tabular*}{\linewidth}{@{\extracolsep{\fill}}llrrrr@{}}
\toprule Dataset & Views & AP & Small AP & Gain & 95\% CI\\\midrule
VOC & Global & 74.42 & 49.76 & --- & ---\\
VOC & Four corners & 78.83 & 64.91 & --- & ---\\
VOC & Nine windows & 81.46 & 68.82 & 7.04 & $[6.45,7.64]$\\
ADE20K & Global & 41.22 & 27.92 & --- & ---\\
ADE20K & Four corners & 48.07 & 37.94 & --- & ---\\
ADE20K & Nine windows & 49.97 & 39.90 & 8.75 & $[8.12,9.38]$\\

\bottomrule\end{tabular*}
\end{table}

To evaluate support annotation sensitivity, we erode or dilate masks while retaining support identities and query observations. The diagnostic uses 400 class-balanced queries each from VOC and COCO and averages metrics over five banks. Perturbations of up to three canvas pixels change AP by at most 0.20 points on VOC and 0.32 on COCO (Table~\ref{tab:boundary}). PCLM is stable under these mask-boundary perturbations, while the support-budget study shows that additional examples can still improve the prototypes.

\begin{table}[tbp]
\def\baselinestretch{1}\selectfont
\centering
\normalsize\renewcommand{\arraystretch}{1.10}
\caption{Support-boundary diagnostic on 400 class-balanced queries per dataset (AP, \%), averaged over five support banks. Radii are canvas pixels; erosion retains a foreground anchor. All paths use the fixed nine windows.}
\label{tab:boundary}
\setlength{\tabcolsep}{4pt}
\begin{tabular*}{\linewidth}{@{\extracolsep{\fill}}lrr@{}}
\toprule
\thd{Annotation} & \thd{VOC \method} & \thd{COCO \method}\\
\midrule
Clean & 80.32 & 66.58\\
Erode 1 & 80.29 & 66.55\\
Erode 3 & 80.12 & 66.26\\
Dilate 1 & 80.32 & 66.58\\
Dilate 3 & 80.24 & 66.51\\

\bottomrule
\end{tabular*}
\end{table}

\FloatBarrier

\section{Discussion and conclusion}
PCLM combines a category reference with finer observations of a query image. Foreground--background contrast determines how each patch is scored, and local magnification increases the number of patches covering a small target. The shared score allows independently encoded views to be aligned and averaged. On the three scene datasets, factorial experiments show that contrast increases the gain from local observation, including when global and local maps use the same image-coordinate filter.

The resolution study distinguishes two ways to obtain finer features. Re-encoding enlarged crops improves small-target AP over enlarging the same canvas at comparable measured latency. Resizing the original RGB image directly also improves global inference by preserving source detail. These choices complement the support budget: more supports improve the category estimate, while more views increase observation cost. The stored direction keeps the same size and can be reused across view layouts and datasets with mapped categories. This is useful when a category is specified once and localized in many subsequent images.

\subsection{Limitations}
\label{sec:limitations}
\begin{figure}[!tb]
\centering\includegraphics[width=\linewidth]{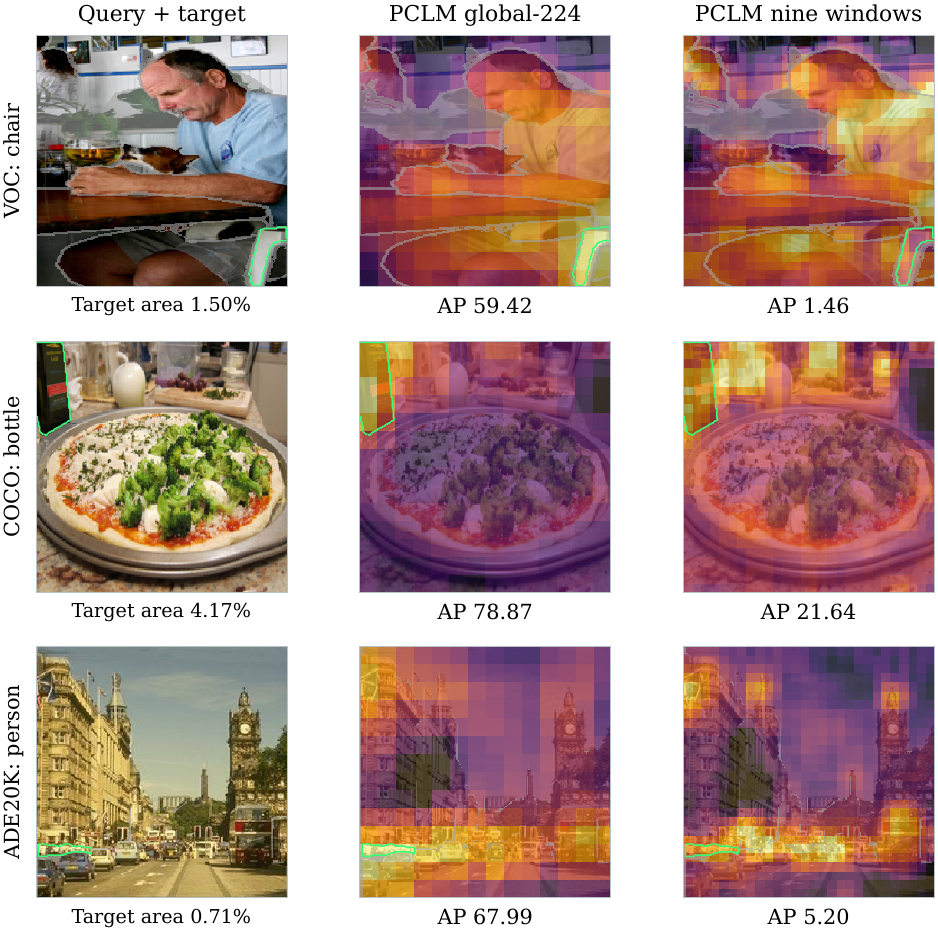}
\caption{Selected failures of local magnification. Global and nine-window PCLM use the same five supports from bank 11. Green contours mark the target and gray shading marks ignored pixels. Heatmaps use per-map 1st--99th percentile display scaling; AP uses original scores over known pixels. Cases were selected from bank-11 small-target queries with global AP of at least 40\%, nine-window AP of at most 35\% and an AP decrease of at least 20 percentage points, followed by visual inspection for one identifiable target per scene dataset.}
\label{fig:failures}
\end{figure}

Local magnification can also increase responses to competing regions. Figure~\ref{fig:failures} shows three such cases with the same support direction for global and local inference. In the VOC chair example, the narrow target at the lower-right edge loses its strong response while the person's head and clothing receive higher scores; AP falls from 59.42 to 1.46. In the COCO bottle example, the left-edge target remains active, but other containers and the counter also receive high scores, reducing AP from 78.87 to 21.64. In the ADE20K person example, nearby road and vehicle regions score above the thin foreground near the lower-left edge; AP falls from 67.99 to 5.20.

These examples show that local re-encoding can change the ranking of target and background regions. Adaptive window placement and query-dependent combinations of global and local maps are promising extensions. Preserving more original-image detail could also improve thin structures and boundaries, which are limited by the initial resizing and the current $8\times8$ output cells.

The current task uses pixel-masked supports and query images known to contain the target. One foreground prototype and one background prototype summarize the appearances in the support set. Multiple prototypes could represent greater variation within a category. Broader category vocabularies and explicit detection of target absence would extend the method to more general localization settings.

\subsection{Conclusion}
PCLM localizes small targets by combining a support-derived foreground--background direction with independently enlarged local views. Each support image contributes equally to the two prototypes, and the resulting 2-KiB direction is shared across query windows. Smoothing, reprojection and coverage averaging turn the patch scores into a dense map using the frozen CLIP encoder.

The experiments identify the contribution of both components. Contrast increases local-observation gains on the three scene datasets. At comparable measured latency, nine views improve small-target AP by 4.51--5.68 percentage points over global enlargement of the same canvas; gains also persist with matched image-coordinate filtering. The same VOC-derived prototypes are reused across mapped categories in all four datasets. Support-budget, backbone and boundary-perturbation studies show how accuracy and stability change with the available examples and encoder.

Across 5,047 small-target queries, PCLM achieves higher mean AP than every evaluated text-conditioned baseline on each dataset, with gains of 5.63--13.23 percentage points over the strongest baseline on each scene dataset. The method provides a compact category representation whose support budget and query observation cost can be chosen separately.

\section*{Data and code availability}
The PCLM implementation, numerical configuration and release checks are publicly available at \url{https://github.com/FourierAI/PCLM}. Benchmark support-image indices, per-query evaluation records, derived benchmark arrays and aggregate benchmark metrics will be released in the same repository upon publication. Original dataset images and annotations and the pretrained CLIP checkpoint are obtained from their respective providers. The dataset papers are cited in Section~\ref{sec:protocol}.

\section*{CRediT authorship contribution statement}
\textbf{Zhipeng Ye:} Conceptualization, Data curation, Formal analysis, Funding acquisition, Investigation, Methodology, Project administration, Resources, Software, Supervision, Validation, Visualization, Writing--original draft, Writing--review and editing.
\textbf{Feng Jiang:} Project administration, Supervision, Writing--review and editing.
\textbf{Qiufeng Wang:} Project administration, Supervision, Writing--review and editing.
\textbf{Hao Li:} Data curation, Software, Validation, Writing--review and editing.

\section*{Acknowledgements}
This work was supported by the Young Scientific and Technological Talent Support Program under the Taizhou Fengcheng Talent Plan. The funder provided financial support only and had no role in the study design, conduct of the experiments, data analysis, manuscript preparation, or decision to submit the article for publication.

\section*{Declaration of competing interests}
The authors declare that they have no known competing financial interests or personal relationships that could have appeared to influence the work reported in this paper.

\section*{Declaration of generative AI and AI-assisted technologies in the manuscript preparation process}
The authors used Codex (OpenAI) to assist with language editing and with implementation, analysis and figure-preparation scripts. They reviewed the outputs, verified the reported results and take responsibility for the article.

\bibliographystyle{elsarticle-num}
\bibliography{references}

\begin{thebibliography}{10}
\expandafter\ifx\csname url\endcsname\relax
  \def\url#1{\texttt{#1}}\fi
\expandafter\ifx\csname urlprefix\endcsname\relax\def\urlprefix{URL }\fi
\expandafter\ifx\csname href\endcsname\relax
  \def\href#1#2{#2} \def\path#1{#1}\fi

\bibitem{radford2021clip}
A.~Radford, J.~W. Kim, C.~Hallacy, A.~Ramesh, G.~Goh, S.~Agarwal, G.~Sastry,
  A.~Askell, P.~Mishkin, J.~Clark, G.~Krueger, I.~Sutskever, Learning
  transferable visual models from natural language supervision, in: Proceedings
  of the 38th International Conference on Machine Learning, Vol. 139 of
  Proceedings of Machine Learning Research, PMLR, 2021, pp. 8748--8763.

\bibitem{agarwal2026cci}
A.~Agarwal, S.~Karanam, V.~Gandhi, Concept regions matter: Benchmarking {CLIP}
  with a new cluster-importance approach, in: Proceedings of the IEEE/CVF
  Conference on Computer Vision and Pattern Recognition, 2026, pp. 2864--2874.

\bibitem{dosovitskiy2021vit}
A.~Dosovitskiy, L.~Beyer, A.~Kolesnikov, D.~Weissenborn, X.~Zhai,
  T.~Unterthiner, M.~Dehghani, M.~Minderer, G.~Heigold, S.~Gelly, J.~Uszkoreit,
  N.~Houlsby, An image is worth 16x16 words: Transformers for image recognition
  at scale, in: International Conference on Learning Representations, 2021.

\bibitem{abnar2020attention}
S.~Abnar, W.~Zuidema, Quantifying attention flow in transformers, in:
  Proceedings of the 58th Annual Meeting of the Association for Computational
  Linguistics, Association for Computational Linguistics, 2020, pp. 4190--4197.
\newblock \href {https://doi.org/10.18653/v1/2020.acl-main.385}
  {\path{doi:10.18653/v1/2020.acl-main.385}}.

\bibitem{chefer2021transformer}
H.~Chefer, S.~Gur, L.~Wolf, Transformer interpretability beyond attention
  visualization, in: Proceedings of the IEEE/CVF Conference on Computer Vision
  and Pattern Recognition, 2021, pp. 782--791.
\newblock \href {https://doi.org/10.1109/CVPR46437.2021.00084}
  {\path{doi:10.1109/CVPR46437.2021.00084}}.

\bibitem{selvaraju2017gradcam}
R.~R. Selvaraju, M.~Cogswell, A.~Das, R.~Vedantam, D.~Parikh, D.~Batra,
  {Grad-CAM}: Visual explanations from deep networks via gradient-based
  localization, International Journal of Computer Vision 128~(2) (2020)
  336--359.
\newblock \href {https://doi.org/10.1007/s11263-019-01228-7}
  {\path{doi:10.1007/s11263-019-01228-7}}.

\bibitem{hsia2022clipcam}
H.-A. Hsia, C.-H. Lin, B.-H. Kung, J.-T. Chen, D.~S. Tan, J.-C. Chen, K.-L.
  Hua, {CLIPCAM}: A simple baseline for zero-shot text-guided object and action
  localization, in: Proceedings of the IEEE International Conference on
  Acoustics, Speech and Signal Processing, 2022, pp. 4453--4457.
\newblock \href {https://doi.org/10.1109/ICASSP43922.2022.9747841}
  {\path{doi:10.1109/ICASSP43922.2022.9747841}}.

\bibitem{zhao2026gradeclip}
C.~Zhao, K.~Wang, J.~H. Hsiao, A.~B. Chan, {Grad-ECLIP}: Gradient-based visual
  and textual explanations for {CLIP}, IEEE Transactions on Pattern Analysis
  and Machine Intelligence (2026) 1--18, early access.
\newblock \href {https://doi.org/10.1109/TPAMI.2026.3714444}
  {\path{doi:10.1109/TPAMI.2026.3714444}}.

\bibitem{bousselham2025legrad}
W.~Bousselham, A.~Boggust, S.~Chaybouti, H.~Strobelt, H.~Kuehne, {LeGrad}: An
  explainability method for vision transformers via feature formation
  sensitivity, in: Proceedings of the IEEE/CVF International Conference on
  Computer Vision, 2025, pp. 20336--20345.
\newblock \href {https://doi.org/10.1109/ICCV51701.2025.01891}
  {\path{doi:10.1109/ICCV51701.2025.01891}}.

\bibitem{chen2022gscorecam}
P.~Chen, Q.~Li, S.~Biaz, T.~Bui, A.~Nguyen, {gScoreCAM}: What objects is {CLIP}
  looking at?, in: Computer Vision -- ACCV 2022, Vol. 13844 of Lecture Notes in
  Computer Science, Springer, 2023, pp. 588--604.
\newblock \href {https://doi.org/10.1007/978-3-031-26316-3_35}
  {\path{doi:10.1007/978-3-031-26316-3_35}}.

\bibitem{li2025clipsurgery}
Y.~Li, H.~Wang, Y.~Duan, J.~Zhang, X.~Li, A closer look at the explainability
  of contrastive language-image pre-training, Pattern Recognition 162 (2025)
  111409.
\newblock \href {https://doi.org/10.1016/j.patcog.2025.111409}
  {\path{doi:10.1016/j.patcog.2025.111409}}.

\bibitem{zhang2023wsfl}
C.-L. Zhang, Y.~Li, J.~Wu, Weakly supervised foreground learning for weakly
  supervised localization and detection, Pattern Recognition 137 (2023) 109279.
\newblock \href {https://doi.org/10.1016/j.patcog.2022.109279}
  {\path{doi:10.1016/j.patcog.2022.109279}}.

\bibitem{cao2025secm}
Y.~Cao, Y.~Su, W.~Wang, Y.~Liu, Q.~Wu, Semantic-constraint matching for
  transformer-based weakly supervised object localization, Pattern Recognition
  158 (2025) 110971.
\newblock \href {https://doi.org/10.1016/j.patcog.2024.110971}
  {\path{doi:10.1016/j.patcog.2024.110971}}.

\bibitem{lee2024inference}
S.~Lee, C.~Mun, Y.~Uh, J.~Choe, H.~Byun, Discovering an inference recipe for
  weakly-supervised object localization, Pattern Recognition 156 (2024) 110838.
\newblock \href {https://doi.org/10.1016/j.patcog.2024.110838}
  {\path{doi:10.1016/j.patcog.2024.110838}}.

\bibitem{vinyals2016matching}
O.~Vinyals, C.~Blundell, T.~Lillicrap, K.~Kavukcuoglu, D.~Wierstra, Matching
  networks for one shot learning, in: Advances in Neural Information Processing
  Systems, Vol.~29, 2016, pp. 3630--3638.

\bibitem{snell2017prototypical}
J.~Snell, K.~Swersky, R.~S. Zemel, Prototypical networks for few-shot learning,
  in: Advances in Neural Information Processing Systems, Vol.~30, Curran
  Associates, Inc., 2017, pp. 4077--4087.

\bibitem{sung2018relation}
F.~Sung, Y.~Yang, L.~Zhang, T.~Xiang, P.~H.~S. Torr, T.~M. Hospedales, Learning
  to compare: Relation network for few-shot learning, in: Proceedings of the
  IEEE Conference on Computer Vision and Pattern Recognition, 2018, pp.
  1199--1208.

\bibitem{zhang2022tip}
R.~Zhang, W.~Zhang, R.~Fang, P.~Gao, K.~Li, J.~Dai, Y.~Qiao, H.~Li,
  {Tip-Adapter}: Training-free adaption of {CLIP} for few-shot classification,
  in: Computer Vision -- ECCV 2022, Vol. 13695 of Lecture Notes in Computer
  Science, Springer, 2022, pp. 493--510.
\newblock \href {https://doi.org/10.1007/978-3-031-19833-5_29}
  {\path{doi:10.1007/978-3-031-19833-5_29}}.

\bibitem{menon2023description}
S.~Menon, C.~Vondrick, Visual classification via description from large
  language models, in: International Conference on Learning Representations,
  2023.

\bibitem{ye2026idea}
Z.~Ye, F.~Jiang, Q.~Wang, K.~Huang, J.~Huang, {IDEA}: Image description
  enhanced {CLIP}-adapter for image classification, Pattern Recognition 171
  (2026) 112224.
\newblock \href {https://doi.org/10.1016/j.patcog.2025.112224}
  {\path{doi:10.1016/j.patcog.2025.112224}}.

\bibitem{zhou2022coop}
K.~Zhou, J.~Yang, C.~C. Loy, Z.~Liu, Learning to prompt for vision-language
  models, International Journal of Computer Vision 130~(9) (2022) 2337--2348.
\newblock \href {https://doi.org/10.1007/s11263-022-01653-1}
  {\path{doi:10.1007/s11263-022-01653-1}}.

\bibitem{zhou2022cocoop}
K.~Zhou, J.~Yang, C.~C. Loy, Z.~Liu, Conditional prompt learning for
  vision-language models, in: Proceedings of the IEEE/CVF Conference on
  Computer Vision and Pattern Recognition, 2022, pp. 16816--16825.

\bibitem{khattak2023maple}
M.~U. Khattak, H.~Rasheed, M.~Maaz, S.~Khan, F.~S. Khan, {MaPLe}: Multi-modal
  prompt learning, in: Proceedings of the IEEE/CVF Conference on Computer
  Vision and Pattern Recognition, 2023, pp. 19113--19122.

\bibitem{gao2024clipadapter}
P.~Gao, S.~Geng, R.~Zhang, T.~Ma, R.~Fang, Y.~Zhang, H.~Li, Y.~Qiao,
  {CLIP-Adapter}: Better vision-language models with feature adapters,
  International Journal of Computer Vision 132~(2) (2024) 581--595.
\newblock \href {https://doi.org/10.1007/s11263-023-01891-x}
  {\path{doi:10.1007/s11263-023-01891-x}}.

\bibitem{lin2017fpn}
T.-Y. Lin, P.~Doll{\'a}r, R.~Girshick, K.~He, B.~Hariharan, S.~Belongie,
  Feature pyramid networks for object detection, in: Proceedings of the IEEE
  Conference on Computer Vision and Pattern Recognition, 2017, pp. 2117--2125.
\newblock \href {https://doi.org/10.1109/CVPR.2017.106}
  {\path{doi:10.1109/CVPR.2017.106}}.

\bibitem{fu2024featup}
S.~Fu, M.~Hamilton, L.~E. Brandt, A.~Feldmann, Z.~Zhang, W.~T. Freeman,
  {FeatUp}: A model-agnostic framework for features at any resolution, in: The
  Twelfth International Conference on Learning Representations, 2024.

\bibitem{shi2025scales}
B.~Shi, Z.~Wu, M.~Mao, X.~Wang, T.~Darrell, When do we not need larger vision
  models?, in: Computer Vision -- ECCV 2024, Vol. 15066 of Lecture Notes in
  Computer Science, Springer, 2025, pp. 444--462.
\newblock \href {https://doi.org/10.1007/978-3-031-73242-3_25}
  {\path{doi:10.1007/978-3-031-73242-3_25}}.

\bibitem{akyon2022sahi}
F.~C. Akyon, S.~O. Altinuc, A.~Temizel, Slicing aided hyper inference and
  fine-tuning for small object detection, in: Proceedings of the IEEE
  International Conference on Image Processing, 2022, pp. 966--970.
\newblock \href {https://doi.org/10.1109/ICIP46576.2022.9897990}
  {\path{doi:10.1109/ICIP46576.2022.9897990}}.

\bibitem{everingham2010voc}
M.~Everingham, L.~Van~Gool, C.~K.~I. Williams, J.~Winn, A.~Zisserman, The
  pascal visual object classes ({VOC}) challenge, International Journal of
  Computer Vision 88~(2) (2010) 303--338.
\newblock \href {https://doi.org/10.1007/s11263-009-0275-4}
  {\path{doi:10.1007/s11263-009-0275-4}}.

\bibitem{everingham2015retrospective}
M.~Everingham, S.~M.~A. Eslami, L.~Van~Gool, C.~K.~I. Williams, J.~Winn,
  A.~Zisserman, The {PASCAL} visual object classes challenge: A retrospective,
  International Journal of Computer Vision 111~(1) (2015) 98--136.
\newblock \href {https://doi.org/10.1007/s11263-014-0733-5}
  {\path{doi:10.1007/s11263-014-0733-5}}.

\bibitem{lin2014coco}
T.-Y. Lin, M.~Maire, S.~Belongie, J.~Hays, P.~Perona, D.~Ramanan,
  P.~Doll{\'a}r, C.~L. Zitnick, Microsoft {COCO}: Common objects in context,
  in: European Conference on Computer Vision, Vol. 8693 of Lecture Notes in
  Computer Science, Springer, 2014, pp. 740--755.
\newblock \href {https://doi.org/10.1007/978-3-319-10602-1_48}
  {\path{doi:10.1007/978-3-319-10602-1_48}}.

\bibitem{zhou2017ade20k}
B.~Zhou, H.~Zhao, X.~Puig, S.~Fidler, A.~Barriuso, A.~Torralba, Scene parsing
  through {ADE20K} dataset, in: Proceedings of the IEEE Conference on Computer
  Vision and Pattern Recognition, 2017, pp. 633--641.
\newblock \href {https://doi.org/10.1109/CVPR.2017.544}
  {\path{doi:10.1109/CVPR.2017.544}}.

\bibitem{zhou2019semantic}
B.~Zhou, H.~Zhao, X.~Puig, T.~Xiao, S.~Fidler, A.~Barriuso, A.~Torralba,
  Semantic understanding of scenes through the {ADE20K} dataset, International
  Journal of Computer Vision 127~(3) (2019) 302--321.
\newblock \href {https://doi.org/10.1007/s11263-018-1140-0}
  {\path{doi:10.1007/s11263-018-1140-0}}.

\bibitem{parkhi2012pets}
O.~M. Parkhi, A.~Vedaldi, A.~Zisserman, C.~V. Jawahar, Cats and dogs, in:
  Proceedings of the IEEE Conference on Computer Vision and Pattern
  Recognition, 2012, pp. 3498--3505.
\newblock \href {https://doi.org/10.1109/CVPR.2012.6248092}
  {\path{doi:10.1109/CVPR.2012.6248092}}.

\bibitem{efron1994bootstrap}
B.~Efron, R.~J. Tibshirani, An Introduction to the Bootstrap, Vol.~57 of
  Monographs on Statistics and Applied Probability, Chapman and Hall, New York,
  1993.

\end{thebibliography}
\end{document}